\documentclass[conference]{IEEEtran}
\IEEEoverridecommandlockouts
\usepackage{cite}
\usepackage{amsmath,amssymb,amsfonts}
\usepackage{algorithmic}
\usepackage{graphicx}
\usepackage{lipsum}
\usepackage{textcomp}
\usepackage{xcolor}
\usepackage{multirow}
\usepackage{booktabs}
\usepackage{float}
\usepackage{array}

\def\BibTeX{{\rm B\kern-.05em{\sc i\kern-.025em b}\kern-.08em
    T\kern-.1667em\lower.7ex\hbox{E}\kern-.125emX}}
\begin{document}

\title{Brainchop: Next Generation Web-Based Neuroimaging Application

}

\author{\IEEEauthorblockN{ Mohamed Masoud}
\IEEEauthorblockA{\textit{TReNDS Center} \\
\textit{Georgia State University}\\
Atlanta, United States of America \\
mmasoud1@gsu.edu}
\and
\IEEEauthorblockN{Pratyush Reddy}
\IEEEauthorblockA{\textit{TReNDS Center} \\
\textit{Georgia State University}\\
Atlanta, United States of America \\
pgaggenapalli1@student.gsu.edu}
\and
\IEEEauthorblockN{Farfalla Hu}
\IEEEauthorblockA{\textit{TReNDS Center} \\
\textit{Georgia State University}\\
Atlanta, United States of America \\
dhu3@student.gsu.edu}
\and

\IEEEauthorblockN{Sergey Plis}
\IEEEauthorblockA{\centerline{Dept. of Computer Science} \\
\textit{Georgia State University}\\
Atlanta, United States of America \\
splis@gsu.edu}
\vspace{-3em}  
}

\maketitle

\begin{abstract}
Performing volumetric image processing directly within the browser, particularly with medical data, presents unprecedented challenges compared to conventional backend tools. These challenges arise from limitations inherent in browser environments, such as constrained computational resources and the availability of frontend machine learning libraries. Consequently, there is a shortage of neuroimaging frontend tools capable of providing comprehensive end-to-end solutions for whole brain preprocessing and segmentation while preserving end-user data privacy and residency. In light of this context, we introduce Brainchop (http://www.brainchop.org) as a groundbreaking in-browser neuroimaging tool that enables volumetric analysis of structural MRI using pre-trained full-brain deep learning models, all without requiring technical expertise or intricate setup procedures. Beyond its commitment to data privacy, this frontend tool offers multiple features, including scalability, low latency, user-friendly operation, cross-platform compatibility, and enhanced accessibility. This paper outlines the processing pipeline of Brainchop and evaluates the performance of models across various software and hardware configurations. The results demonstrate the practicality of client-side processing for volumetric data, owing to the robust MeshNet architecture, even within the resource-constrained environment of web browsers.
\end{abstract}

\begin{IEEEkeywords}
Volumetric segmentation, MeshNet,  MRI, 3D dilated CNN.
\vspace{-1em} 
\end{IEEEkeywords}

\section{Introduction}
Extracting brain tissue from structural Magnetic Resonance Imaging (MRI) volumes and subsequent segmentation into gray and white matter, or more elaborate brain atlases, is essential to brain imaging analysis pipelines. Fostering the advancement of automatic medical image segmentation is vital to improving the precision and efficacy of clinical diagnoses. Clinical applications such as surgical planning, detection of brain atrophy, and visualization of anatomical structures heavily rely on MRI segmentation. However, for numerous researchers and radiologists, especially those in developing countries, establishing neuroimaging pipelines poses technological barriers. Offering these pipelines through browser-based platforms can contribute to democratizing computational approaches in these contexts.
Nevertheless, leveraging the browser for neuroimaging applications entails confronting multiple challenges, including limitations in memory and computational resource management. Consequently, there exists a shortage of web-based neuroimaging tools capable of providing fast and reliable volumetric brain segmentation while maintaining strict end-user data privacy and residency. Despite the better accuracy and training convergence achieved by volumetric segmentation models compared to sub-volume and 2D segmentation models[12], the existing tools for segmentation in the browser either lack volumetric inference or need backend support. While backend-based medical image applications raise privacy issues surrounding hosting or accessing raw user data, hybrid methods involving distributed deep learning processing between the client and the cloud have not yielded practical solutions regarding medical data privacy, which flags the importance of investigating in-browser tools as potential alternatives capable of resolving the data privacy issue with low latency,  as it enables the direct execution of deep learning models on the client side. By "client-side" and "browser inference," we refer to the entirety of the computational task being executed on the user side, eliminating the need to transfer data to remote servers for processing. However, despite the recent advancements in deep learning frameworks in JavaScript, such as TensorFlow.js and its model deployment and conversion techniques, tasks such as volumetric segmentation for MRI images, which typically entail substantial computational workloads, remain challenging for inference within the browser resource-constrained environment.

This work represents our innovative online pipeline Brainchop (http://www.brainchop.org), designed to facilitate brain image processing and segmentation. Notably, Brainchop stands out as the first in-browser tool that enables scientists and clinicians to perform volumetric analysis of structural MRI utilizing pre-trained deep learning models, all without necessitating technical proficiency or the setup of AI solutions. It delivers valuable attributes, including data privacy, enhanced accessibility, scalability, low latency, user-friendly operation, elimination of installation requirements, and seamless cross-platform functionality while preserving MRI data privacy. 

Building upon our previous work [1], this paper delves into a meticulous analysis of Brainchop's performance characteristics across various models and resource configurations.

\begin{figure}[t]
    \centering
    \includegraphics[width=9cm]{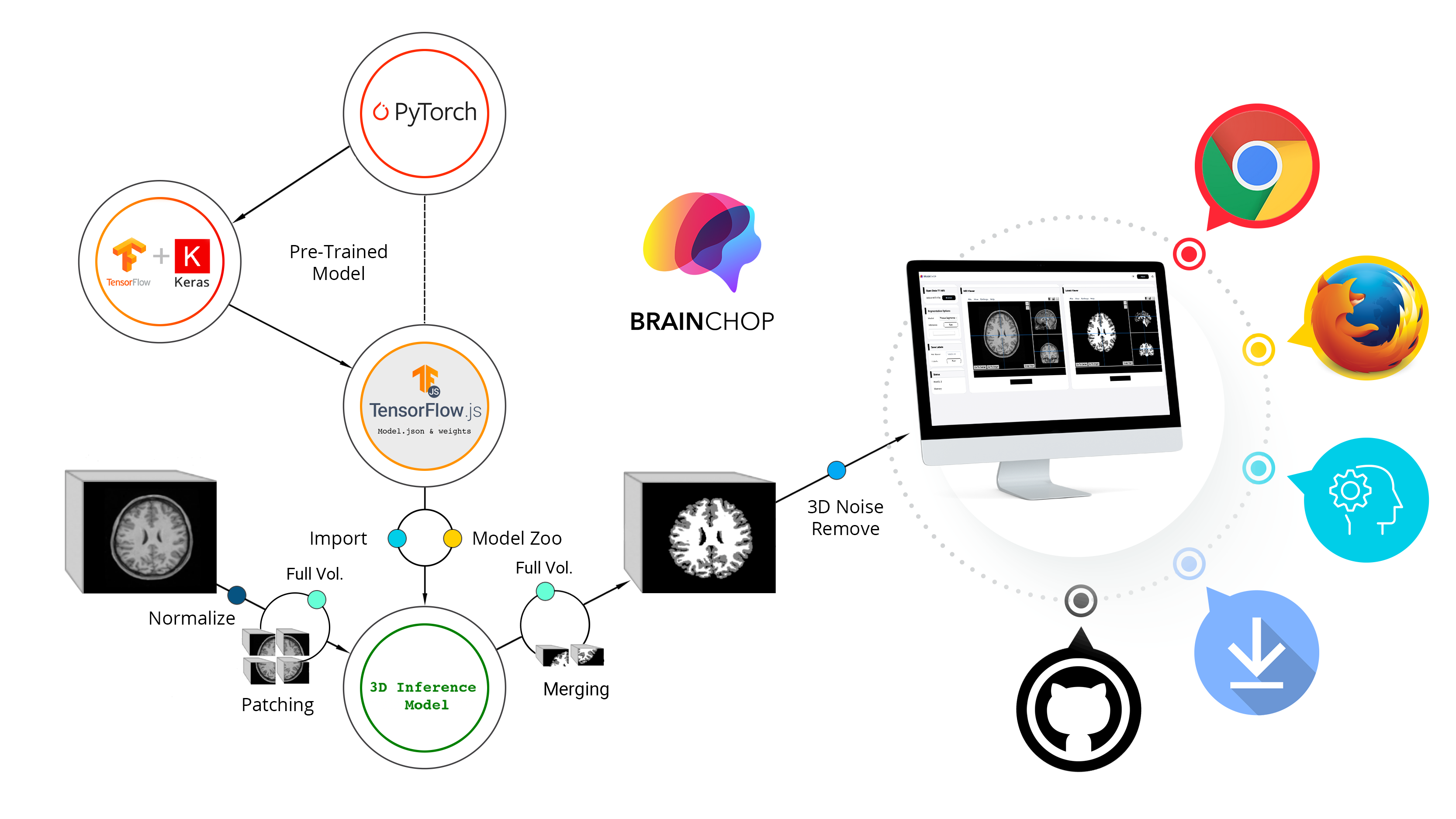}
    \caption{The Brainchop high-level architecture allows converting pre-trained models in PyTorch and Keras to Tensorflow.js, enabling their importation into the Brainchop models list. The input MRI data can be handled in two ways: it can be passed as a complete volume to the inference model or divided into subvolumes to overcome memory limitations in web browsers. In the latter case, the inference output is generated by merging the subvolumes. It is important to note that the inference process may introduce 3D noisy regions, which can be attributed to biases, variances, and irreducible errors such as data noise. We have developed a 3D connected components algorithm to address this issue that effectively filters out these noisy regions.}
    \label{fig:BrainchopPipleline}
  \vspace{-1em}    
\end{figure}

\section{METHODOLOGY}
Brainchop, an open-source front-end application, is developed to enable MRI data resampling, preprocessing, segmentation, and postprocessing in the browser (Fig.1). Notably, it can process whole brain volume in a single pass for segmentation by using the lightweight and reliable MeshNet model [3]. Meshnet, as a variant of dilated convolutions [4], incorporates a volumetric option that enhances the accuracy of MRI inference while maintaining modest computational requirements. The MeshNet segmentation models are trained in Pytorch using the Human Connectome Project (HCP) dataset [5] and a processed FreeSurfer segmentation. Subsequently, the pre-trained models are converted to TensorFlow.js [6] to enable in-browser inference. 

Brainchop is designed to support T1-weighted MRI volume segmentation, with input expected in Nifti format [7]. As a preprocessing step to obtain accurate results, the T1 image should be shaped to $256^{3}$ and resampled to 1 mm isotropic voxels. This preprocessing task can be conveniently performed with Brainchop using mriconvert.js, which employs Pyodide [8] to deploy the "conform" function from FastSurfer [9]. This function is responsible for reshaping, scaling, and resampling the raw T1 image data. Additionally, Brainchop integrates standard medical image preprocessing techniques to eliminate noisy voxels from the input and enhance MRI volume intensities, thus facilitating efficient in-browser inference with optimal results.

\begin{table}[t]
  \caption{Hyperparameters for Typical GWM MeshNet Model (Stride 1)}
  \label{tab:hyperparameters}
  \centering
   \addtolength{\tabcolsep}{-2pt} 
  \setlength\extrarowheight{-4pt}
  \begin{tabular}{ccccccc}
    \toprule
    \multirow{2}{*}{Layer} & \multirow{2}{*}{Type} & \multirow{2}{*}{InCh} & \multirow{2}{*}{OutCh} & \multirow{2}{*}{Kernel} & \multirow{2}{*}{Padding} & \multirow{2}{*}{Dilation}\\
     & & & & & \\
    \midrule
    1 & Conv3d & 1 & 5 & $3^{3}$ & (1, 1, 1) & (1, 1, 1) \\
     & BN3d & 5 & 5 & - & - & - \\
     & ReLU & 5 & 5 & - & - & -\\
     & Dropout3d & 5 & 5 & - & - & - \\
     \midrule
    2 & Conv3d & 5 & 5 & $3^{3}$ & (2, 2, 2)  &(2, 2, 2) \\
     & BN3d & 5 & 5 & - & - & - \\
     & ReLU & 5 & 5 & - & - & -\\
     & Dropout3d & 5 & 5 & - & - & - \\
     \midrule
    3 & Conv3d & 5 & 5 & $3^{3}$ & (4, 4, 4)  &(4, 4, 4) \\
     & BN3d & 5 & 5 & - & - & - \\
     & ReLU & 5 & 5 & - & - & -\\
     & Dropout3d & 5 & 5 & - & - & - \\
     \midrule
    4 & Conv3d & 5 & 5 & $3^{3}$ & (8, 8, 8) & (8, 8, 8) \\
     & BN3d & 5 & 5 & - & - & - \\
     & ReLU & 5 & 5 & - & - & -\\
     & Dropout3d & 5 & 5 & - & - & - \\
     \midrule
    5 & Conv3d & 5 & 5 & $3^{3}$ & (16, 16, 16)  &(16, 16, 16) \\
     & BN3d & 5 & 5 & - & - & - \\
     & ReLU & 5 & 5 & - & - & -\\
     & Dropout3d & 5 & 5 & - & - & - \\
     \midrule
    6 & Conv3d & 5 & 5 & $3^{3}$ & (8, 8, 8) &(8, 8, 8) \\
     & BN3d & 5 & 5 & - & - & - \\
     & ReLU & 5 & 5 & - & - & -\\
     & Dropout3d & 5 & 5 & - & - & - \\
     \midrule
    7 & Conv3d & 5 & 5 & $3^{3}$ & (4, 4, 4)  &(4, 4, 4) \\
     & BN3d & 5 & 5 & - & - & - \\
     & ReLU & 5 & 5 & - & - & -\\
     & Dropout3d & 5 & 5 & - & - & - \\
     \midrule
    8 & Conv3d & 5 & 5 & $3^{3}$ & (2, 2, 2) &(2, 2, 2) \\
     & BN3d & 5 & 5 & - & - & - \\
     & ReLU & 5 & 5 & - & - & -\\
     & Dropout3d & 5 & 5 & - & - & - \\
     \midrule
    9 & Conv3d & 5 & 5 & $3^{3}$ & (1, 1, 1) &(1, 1, 1) \\
     & BN3d & 5 & 5 & - & - & - \\
     & ReLU & 5 & 5 & - & - & -\\
     & Dropout3d & 5 & 5 & - & - & - \\
     \midrule
    10 & Conv3d & 5 & 3 & $1$ & - &(1, 1, 1) \\
    \bottomrule
    \vspace{-3em}  
  \end{tabular}
\end{table}

To ensure the quality of the segmentation output, a 3D connected components algorithm is implemented within the pipeline postprocessing stage to filter out noisy voxels and regions resulting from the inference stage. Both the input MRI data and the resultant segmentation can be viewed using Papaya [10]. Additionally,  Brainchop incorporates a 3D volume rendering functionality powered by Three.js [11] library, enabling users to subjectively verify the accuracy of volumetric segmentation and enhance their visualization experience. All these functions are provided in a user-friendly interface that features simplicity, privacy preservation, and efficiency.
\vspace{-1em}  

\section{Model Training}
\vspace{-1em}  
MeshNet is a feed-forward 3D convolutional neural network with dilated kernels. We trained a model of nine layers to segment brain tissue into  Gray White Matter (GWM) labels, as illustrated in Fig. 2. Each layer incorporates 3D dilated convolutions with a specific padding setting and dilation factor carefully chosen to modify the receptive field for best capturing a broader range of contextual information from the input data without significantly increasing the number of network parameters. The volumetric dilated convolution can be formulated as follows:
\begin{equation}
(k*_lf)_{(x,y,z)}\!=\!\sum_{\bar{x}=-a}^a \sum_{\bar{y}=-b}^b \sum_{\bar{z}=-c}^c k(\bar{x},\bar{y},\bar{z})f(x-l\bar{x},y-l\bar{y},z-l\bar{z})
\label{eq:dil3dconv}
\end{equation}

Where $a$, $b$, $c$ are kernel $k$ bounds on $x$, $y$ and $z$ axis, respectively, and $l$ is the dilation factor specifying gaps between the kernel elements for configuration of the receptive field.

additionally, to enhance the performance and robustness of the
network, each layer incorporates additional techniques, such
as 3D batch normalization, ReLU activation, and 3D dropout
regularization, as outlined in Table-I.

\begin{figure}[b]
    \centering
    \includegraphics[width=9cm]{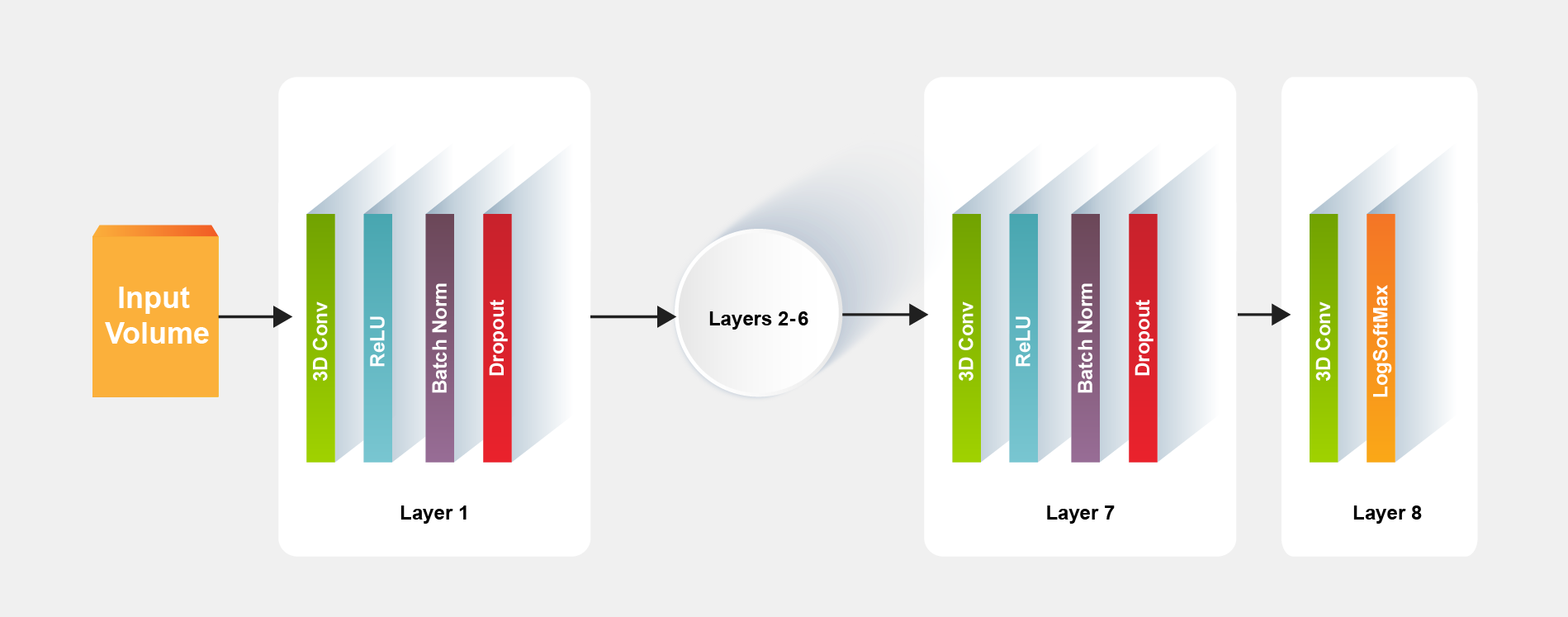}
    \caption{MeshNet model architecture.}
    \label{fig:DL_Arch}
\end{figure}


More information about the MeshNet training tutorial is given in Section-V. The tutorial shows the inference of both Full-Volume and Sub-Volumes and an implementation of a custom DataLoader to handle large MRI volumes and streamline the training of the MeshNet networks.

\begin{table}[t]
  \caption{MeshNet performance.}
  \label{tab:MeshnetPerformance.}
  \centering
  \setlength\extrarowheight{-4pt}
  \begin{tabular}{ccc}
    \toprule
    \multirow{2}{*}{Model} & \multirow{2}{*}{Model Size} & \multirow{2}{*}{Macro Dice}  \\
     & &  \\
    \midrule
    MeshNet GWM (Sub Volume Version)  &  0.89 mb      &  0.96  \\
    MeshNet GWM (Full Volume Version) &  0.022 mb   &  0.96 \\
    U-Net GWM (Sub Volume Version)    &  288 mb     &  0.96 \\
     \midrule
    \bottomrule
  \end{tabular}
  \vspace{-1em}  
\end{table}

\begin{table*}[t]
  \centering
  \setlength\extrarowheight{-3pt}
  \begin{tabular}{*{10}{c}}
    \hline \\ [-1ex] 
     & {\bf Country} & {\bf City} &   {\bf Date} &  {\bf Browser Ver} & {\bf OS} & 
       {\bf Model} & {\bf GPU Card} & {\bf GPU Vendor} & {\bf Status}  \\ [1ex]
     \hline \\ [-1ex]
    {\bf Unique} & 70 & 388 & 276 & 67 & 4 & 12 & 180 & 14 & 2    \\ [1ex]

    {\bf Top}   & United States & Gumi & 5/21/2023 & Chrome 110 & Windows & Full Brain GWM(light) & Apple M1 & Intel & OK  \\ [1ex]
  
    {\bf Top-Freq}  & 428 & 55 & 34 & 125 & 782 & 510 & 109 & 390 & 1095   \\ [1ex]
    \hline \\
  \end{tabular}
  \caption{Brainchop selected categorical telemetry data from {\bf 1336} samples till May 2023 }
\end{table*}

\subsection{DataLoader Implementation}
We implemented a custom DataLoader using the DataLoaderClass to facilitate data loading and preprocessing. This DataLoader effectively handles the following:
\vspace{0.5em}  
\subsubsection{Data Loading}
Using the nibabel library [13], it loads the corresponding images and labels. This step ensures seamless integration of the dataset into our experiments.
\vspace{0.5em}  
\subsubsection{Subvolumes Generation (optional)}
Leveraging the CubeDivider class, the DataLoader partitions the loaded images and corresponding labels into sub-cubes. This subdivision optimizes memory utilization throughout the training process.
\vspace{-0.5em}  
\subsubsection{Data Preparation}
The DataLoader reshapes the subvolumes to match the desired input size for our neural network. Additionally, the labels are converted to one-hot encoding, simplifying multi-class classification tasks.
\vspace{\baselineskip}
\subsubsection{Data Batching}
To optimize training, the DataLoader organizes the preprocessed subvolumes into batches to accelerate the training process.
\subsection{Metrics}
We use Dice metrics and Cross Entropy loss for model training in our experiment to increase the model efficiency
\vspace{-0.5em} 
\subsubsection{Dice Metrics}
During the model training, Dice scores for label maps are computed using binary masks and logical operations to quantify the intersection of pixels between the label maps. The Dice score is calculated using the formula below:
\begin{equation} \label{eq:dice}
  DICE = \frac{2 |X \cap Y|}{|X| + |Y|}
\end{equation}

Where $X$ is the predicted mask and $Y$ is the ground truth one. The Dice score quantifies the degree of overlap between the predicted and true labels, assigning a value of 1 when the segmentation results are identical.
\vspace{1em} 
\subsubsection{CrossEntropy Loss}
It is a commonly used loss function in machine learning for tasks such as classification. It measures the dissimilarity between predicted probabilities and true labels, measuring how well a model performs.
\begin{equation}  
Cross\: Entropy\: Loss = -\sum(y \cdot \log(p))
\end{equation}
\begin{itemize}
  \item $y$ is the true label or target value.
  \item $p$ is the predicted probability assigned by the model to the corresponding class or category
\end{itemize}
\vspace{0.0em}  
The advantage of MeshNet architecture is its compact size and a minimal number of parameters, making it suits for in-browser inference. Meanwhile, the model can still achieve a competitive Dice score compared to the classical U-Net model, as shown in Table-2.

\begin{table*}[t]
  \centering
  \addtolength{\tabcolsep}{-3pt} 
  \begin{tabular}{*{9}{c}}
    \hline \\ [-1ex]
    {\bf Model} & {\bf Layers} & {\bf Parameters} &   {\bf Preprocessing} & {\bf Cropping} & {\bf Inference} & {\bf Merging} & {\bf Postprocessing} & {\bf Requested}  \\ [1ex]
      &   &   &  {\bf Avg. Time(S)} &  {\bf Avg. Time(S)}  & {\bf Avg. Time(S)} & {\bf Avg. Time(S)} & {\bf Avg. Time(S)} & {\bf Texture Size}  \\ [1ex]
    \hline \\ [-1ex]
    Compute Brain Mask (FAST) & 20 & 5598 & 1.899036 & - & 7.551620 & - & 18.739939 & 9159  \\ [1ex]
     
     Extract the Brain (FAST) &  20 & 5598 & 2.280892 & - & 7.870467 & - & 15.030079 & 9159  \\ [1ex]
     
    Full Brain GWM (large)  &  18 & 23290 & 1.647580 & - & 14.011876 & - & 13.194727 & 13585  \\ [1ex]
     
    Full Brain GWM (light) & 20 & 5598 & 2.624046 & - & 9.935594 & - & 14.732657 & 9159   \\ [1ex]
     
    Subvolume GWM (failsafe) & 20 & 96078 & 2.697981 & - & 39.773656 & 1.991998 & 12.707437 & 8192  \\ [1ex]
     
    Compute Brain Mask (failsafe) & 16 & 72222  & 2.124019 & - & 27.754568 & 2.161475 & 16.055606 & 8192  \\ [1ex]
     
    Compute Brain Mask (High Acc) & 18 & 23290  & 0.976887 & - & 17.421975 & - & 14.490000 & 13585  \\ [1ex]
     
    Extract the Brain (failsafe) & 16 & 72222  & 1.932426 & - & 30.317263  & 2.163195 & 12.805639 & 8192  \\ [1ex]
       
    Extract the Brain (High Acc) & 18 & 23290  & 1.111473 & - & 8.232091 & - & 11.580500 & 13585  \\ [1ex]
      
    Cortical Atlas 50  & 20 & 27132 & 1.596227 & 27.202827 & 8.525125 & - & 11.803998 & 16384  \\ [1ex]
     
    FS aparc+aseg Atlas 104 (failsafe) & 18 & 86372 &  1.710471 & 29.035294 & 37.802824 & 16.573937 & 19.328529 & 16384  \\ [1ex]
     
    FS aparc+aseg Atlas 104 & 18 & 86372 & 1.340567 & 21.924533 & 13.652400 & - & 11.471533 & 32768  \\ [1ex]
    \hline   \\
  \end{tabular}
  \caption{Brainchop performance per segmentation model.}
  \vspace{-2.5em} 
\end{table*}

\section{Results}
Despite the high diversity of computational resources available on the user side, the overall success rate of Brainchop is around  82\%, as shown in Fig. 3,  and this percentage is expected to increase with the annual advances in computational resources.

\begin{figure}[H]
    \centering
    \includegraphics[width=6cm]{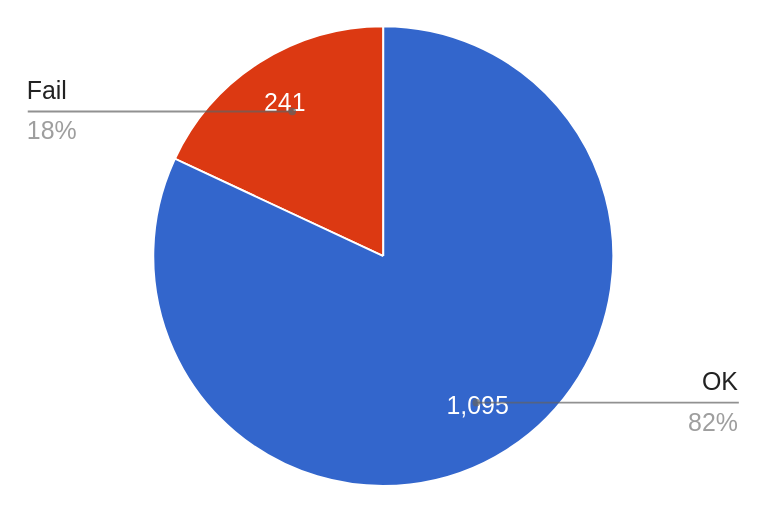}
    \caption{Brainchop shows a success rate of 82\% based on 1336 access instances. }
    \label{fig:BC_successful}
\end{figure}

Multiple volumetric segmentation tasks are available with Brainchop using our pre-trained Pytorch models converted to TensorFlow.js for in-browser inference with WebGL backend. The tasks included brain masking, gray matter white matter (GWM) segmentation, and brain atlas models for 50 cortical regions and 104 cortical and subcortical structures. A list of the models and their performance is given in Table-IV.

By conducting user research and collecting anonymized telemetry data, Brainchop demonstrated a high usability rate among the scientific community, with 1336 hits from its first release in May 2022 till the end of May 2023. A brief description of selected columns of the telemetry data and their unique values are given in Table-III.   A sample of the data is available in the tool's public repository for exploration.

 \begin{figure*}[t]
    \centering
    \includegraphics[scale=0.35]{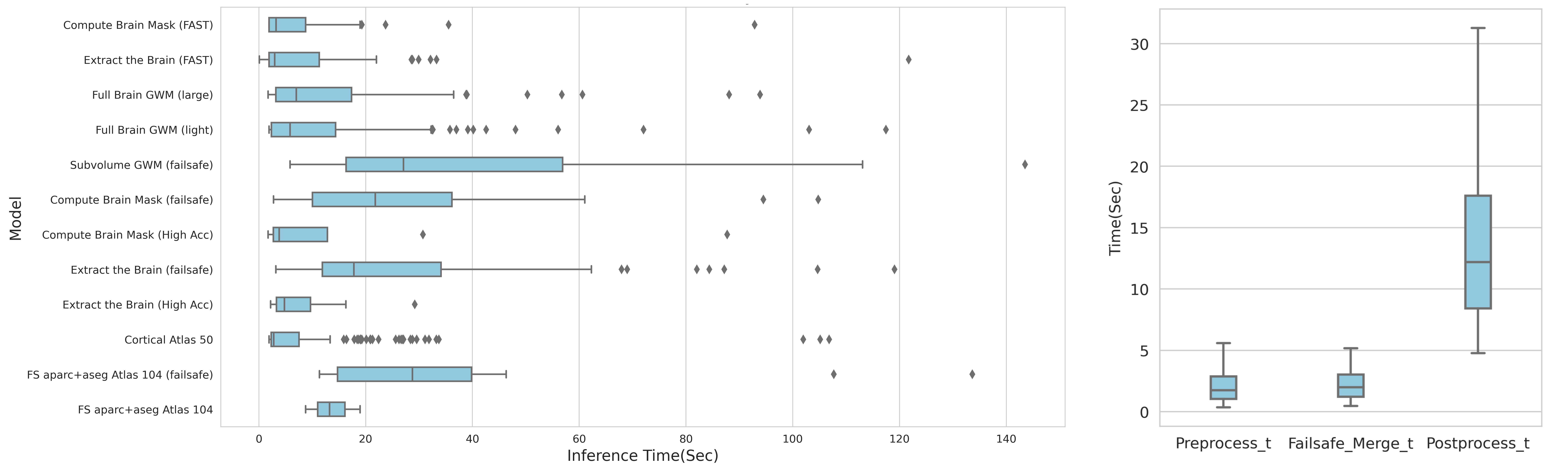}
    \caption{Brainchop overall processing porformance. (Left) Inference performance per model. (Right) The total samples box-plot for the preprocessing, subvolumes merging and postprocessing.  }
\end{figure*}

We conducted a comprehensive analysis of the collected dataset, comprising both categorical and numerical variables, while focusing on analyzing the factors that affect the tool success rate. For exploring the tool Status column as the outcome, we established it as a binary variable indicating whether a tool succeeded or failed during the performed task. As a preprocessed step, the data is cleaned by excluding extreme outliers, and features with correlation coefficients of high similarities  ($Threshold > 0.95$) are pruned, such as those related to the heap size. The selected dataset features (columns) and use cases (rows) are free of missing values, allowing us to proceed with the analysis without the need for imputation. The label encoder is used for categorical data encoding, while the one-hot encoding is utilized with regression models to capture each categorical value's effect independently.

\begin{table}[b]
  \caption{Fail types vs models versions }
  \label{tab:ErrorTypes.}
  \centering
  \setlength\extrarowheight{-4pt}
  \addtolength{\tabcolsep}{-3pt} 
  \begin{tabular}{ccc}
    \toprule
    \multirow{2}{*}{Fail Type} & \multirow{2}{*}{Full Volume} & \multirow{2}{*}{Sub-Vol (Failsafe)}  \\
     & &  \\
    \midrule
    Failed to compile fragment shader  &  174     & 19  \\
    Failed to link vertex and fragment shaders &  33   &  5 \\
    Unable to create WebGL Texture    &  10     &  0 \\
     \midrule
    Total Fails  &  217     & 24  \\     
    OK &  930   &  165 \\    
    \midrule
    Success Rate (\%) &  81.08\%  & 87.3\% \\
    \midrule      

  \end{tabular}
\end{table}

The power analysis of the collected data is performed using the Chi-Square test for independence to determine the significant relation between the selected features and tool status. The overall statistical power of the collected telemetry data was 0.963 for a desired significance level of 0.05, which reflects the adequate sample size of the data to correctly reject the null hypothesis if the alternative hypothesis is true.

\textbf{Statistical analysis:} Statistical significance for null hypothesis testing was defined using $95\%$ confidence intervals ($P<0.05$).

In order to enhance Brainchop's performance, multiple interventions such as patching (sub-volumes) and cropping for input data are applied. The inference models provided include full-volume and sub-volume (Failsafe) models to meet the high diversity of existing computational resources and their possible limitations. The fail status shown in Table-V is mainly caused by limited GPU memory space, as evidenced by the higher success rate of sub-volume models versus full-volume models in Table-V. However, the main drawback of the patching approach (sub-volume models) is its slow inference time, less accuracy, and the overhead cost of the merging step compared to full-volume inference, as shown in Fig. 4.

Estimating the patching effect accurately in the light of causal analysis requires identifying the patching intervention as a treatment and isolating its effect from other potential or significant confounders. To determine the covariates that may confound the relationship between the patching effect and the tool success rate, we conducted a potential confounder analysis using the Chi-Square test for with a significance level 0.05. The list of potential confounders is filtered based on their p-values calculated by the Ordinary Least Squares(OLS)  regression model. The results show the significance of cropping (i.e., Input Shape) in influencing the patching effect, besides the other less significant confounders. To make the patching treatment independent of the cropping confounding variable and estimate its effect, we used regression adjustment that shows a patching effect of 10.4\% on the success rate independent of the cropping effect.

\begin{table}[b]
  \caption{patching and Cropping effects  }
  \label{tab:fullSubEffect.}
  \centering
  \setlength\extrarowheight{-4pt}
  \begin{tabular}{cccc}
    \toprule
    Texture Size       &  Sub-Vol(Failsafe) & Full-Volume & Full-Volume  \\
    Input Crop           &  -      &  -      &   \checkmark   \\
    \midrule 
    Fail               & 7       &  213    &    4 \\     
    OK                 & 148     &  759    &   171 \\    
    \midrule
    Success Rate (\%)  & 95.48\%  & 78.09\%  & 97.71\% \\
    \midrule      

  \end{tabular}
\end{table}

To demonstrate the isolated patching effect, we used the exclusion of samples technique shown in Table-VI to create homogeneous groups without cropping. This allows for a comparison between Sub-Volume and Full-Volume, removing the influence of the cropping effect on the tool success rate.

From Table-VI, the cropping effect with full volume is more significant than the patching effect on the success rate. To validate the result, a multivariable analysis is applied to investigate the effect of the two treatments, cropping and patching, on the tool success rate. By including both treatments simultaneously, we can find their independent effects on the tool success rate while accounting for potential confounding or interactions. The results show the estimated effect of each treatment on the success rate such that the cropping estimated coefficient is 0.0932, indicating that, on average, a one-unit increase in the cropping variable is associated with a 9.32\% increase in the tool success rate, holding patching effect constant. For the patching effect, it shows an estimated coefficient of 5.97\%.

However, the exclusion approach results in a reduced sample size that needs careful consideration to avoid biased estimation or lower statistical power. A more robust technique that avoids such a drawback and, meanwhile, considers the other less significant confounds in our analysis is the Randomized Controlled Trial (RCT). In that technique, a random assignment of the patching treatment is used while ensuring the randomization of other confounds across the treatment group to control the effect of those confounds. In that context, Inverse Probability of Treatment Weighting (IPTW) [14] can help reduce confounding bias in a dataset when estimating causal effects by attempting to mimic the characteristics of a randomized controlled trial.   By reweighting the observations based on the estimated probabilities of treatment assignment, IPTW aims to balance the covariate distributions between treated and control groups, reducing the confounding bias.

 \begin{figure*}[t]
    \centering
    \includegraphics[scale=0.35]{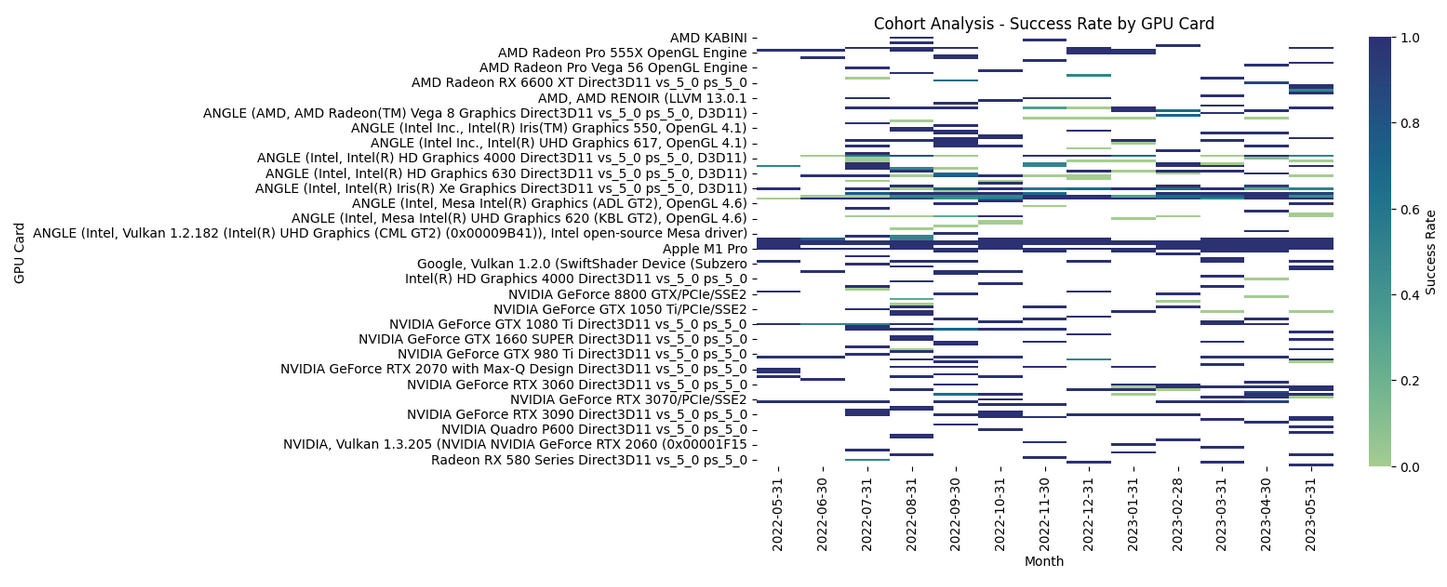}
    \caption{Cohort Analysis - Success Rate by GPU Card Per Month }
\end{figure*}

The Average Treatment Effect  (ATE) on the entire sample can be estimated such that $ATE=p( Outcome=1|do(Treatment=1) - p(Outcome=1|do(Treatment=0)$. In our case, it is the probability of success rate when we apply the treatment (e.g., patching or cropping) versus the probability when we do not.

The estimations of the patching effect using IPTW show an increase in the Brainchop success rate by 6.23\%  due to patching the MRI into subvolumes, an increase in the  inference time by 24.31 seconds, as can also be evident from Fig. 4,  and show almost no change in the postprocessing time with a slight decrease of 0.04 second.

The Atlas models (i.e., 50 and 104 labels) are memory-hungry. Consequently, volumetric cropping is an essential step for the MRI by using the brain masking model, which is applied to exclude the surrounding background from the MRI, resulting in a substantial reduction in volume size and a decrease in the allocated memory, thus helping in making the parcellation possible in the browser.

\begin{table}[t]
  \caption{Input Cropping effect on Full Volume inference}
  \label{tab:CropEffect.}
  \centering
  \setlength\extrarowheight{-4pt}
  \begin{tabular}{cccccc}
    \toprule
    Parameters    & 5598   &  23290  &  27132        & 86372   \\
    Layers        &  20    & 18      &   20          & 18    \\
    Input Crop    &  -     &  -      &   \checkmark  &  \checkmark \\
    \midrule
    Fail          &  135   & 78      &  3            & 1 \\     
    OK            &  644   & 115     &  168          & 3\\    
    \midrule
    Success Rate (\%) &  82.67\%  & 59.56\%  & 98.25\%  & 75.00\%\\
    \midrule      

  \end{tabular}
\end{table}

As presented in Table VII for full volume inference, the Chi-Square test for the Status-Cropping contingency table  indicates a statistically significant association between the success rate of Brainchop and the cropping effect (p-value $2^{-09}$). The statistical power analysis of Table-VII sample size indicates a probability of 99.9\% to reject the null hypothesis correctly.

Estimating the cropping effect using IPTW shows an increase in the Brainchop success rate by 18.12\%  due to cropping the MRI input volume, a decrease in the inference time by 5.26 seconds,  and a decrease in the postprocessing time by 6.83 seconds.

\begin{table}[b]
  \caption{Texture Size effect on Full Volume inference}
  \label{tab:TextureSizeEffect.}
  \centering
  \setlength\extrarowheight{-4pt}
  \begin{tabular}{ccc}
    \toprule
    Texture Size       &  16384  &  32768    \\
    \midrule 
    Fail              & 216      &  1  \\     
    OK                 & 872     &  57 \\    
    \midrule
    Success Rate (\%)  & 80.15\%  & 98.27\% \\
    \midrule      

  \end{tabular}
\end{table}

Table-VIII shows that larger texture sizes can reduce memory fragmentation errors and increase the success rate. Performing the Chi-Square test shows a statistically significant association between the tool success rate and texture size (p-value 0.0024) in full volume inference with a statistical power of 0.934 to reject the null hypothesis correctly.

When increasing the texture size to 32768, the texture size effect on Brainchop performance shows an increase in the tool success rate by 18.13\%, a decrease in the inference time by 2.3 seconds,  and a decrease in the postprocessing time by 5.70 seconds.

Our results also show a marginal rise in the mean heap size and the number of logical CPU cores within the successful instances compared to failed ones, which can be explained as an increase in the browser's capability to handle concurrent tasks more efficiently by using the web workers in parallel with the main browser thread. Such an approach can prevent bottlenecks, enhance asynchronous tasks, and reduce main thread blocking due to resource-intensive computations.

 \begin{figure}[h]
    \centering
    \includegraphics[scale=0.25]{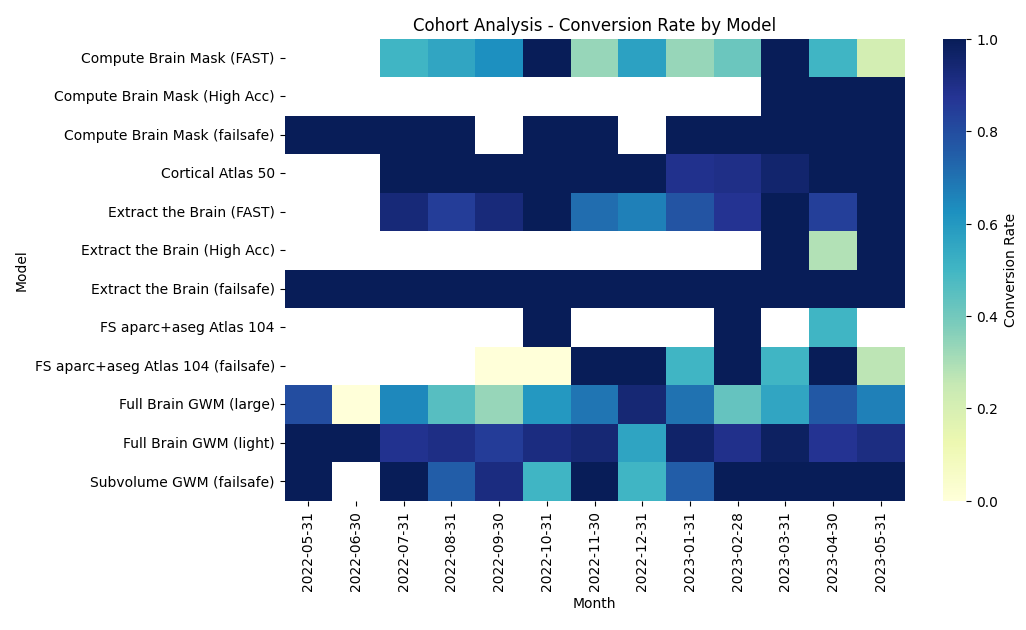}
    \caption{Cohort Analysis - Success Rate by Model  }
\end{figure}

Full volume inference requires careful consideration to retain the efficiency of in-browser processing without memory leaks or loss of the WebGL context. In order to mitigate memory leakage and effectively handle the substantial memory requirements while simultaneously minimizing instances of failure, an inference strategy was adopted. This approach entails the progressive utilization of the MeshNet model on a layer-by-layer basis, coupled with the strategic disposal of the MRI tensor from the preceding layer. This tactic was implemented to alleviate memory-related challenges.

\textbf{Limitations:} Despite the statistical power of the telemetry data, applying stratification analysis may lack sufficient subgroups due to the high diversity of computational resource configuration. The success rate over time by GPU in Fig. 5 and by Models in Fig. 6  are mutually dependent such that a model success rate depends on the GPU in use and vice versa. For the brain masking model(Fast), although having a high success rate for its moderated number of parameters, it only shows an average success rate when used as a pre-model for cropping input data before applying Atlas models, which raises the need for further investigation. 

In general, Brainchop demonstrates a high success rate and processing speed for volumetric segmentation in the browser, with potential for further improvement. The success rate percentage is expected to increase with the continual advances in computational resources supported by a consistent tendency to increase the gap between the successful and failed tasks, as shown in Fig. 7.

\section{Code Availability }

Brainchop source code is publicly available on  GitHub (https://github.com/neuroneural/brainchop). The Pytorch training pipeline is also provided in a Google Colab. A sample of the telemary dataset is accessible with the Wiki step-by-step documentation.


\section{Conclusion}
Through our meticulous analysis, we have unveiled valuable insights into Brainchop's overall performance. Our analysis determined a statistically significant correlation between patching, cropping, texture size,  and both the timing and success rate of Brainchop. Notably, Brainchop has exhibited a high success rate of 82\%. This accomplishment and its potential for further enhancement underscore its promise as a browser-based neuroimaging solution. 
Our findings also highlight the need to refine the cropping techniques for better outcomes. Additionally, a more in-depth exploration into the current limitations of the tool holds the potential to provide further insights, which in turn can inform efforts to optimize Brainchop performance.

In summation, our analysis has not only shed light on the drivers of tool success rates but also provided metrics that can assist frontend tools in performing volumetric segmentation, thus enhancing the user experience significantly while maintaining data privacy.

 \begin{figure}[t]
    \centering
    \includegraphics[scale=0.25]{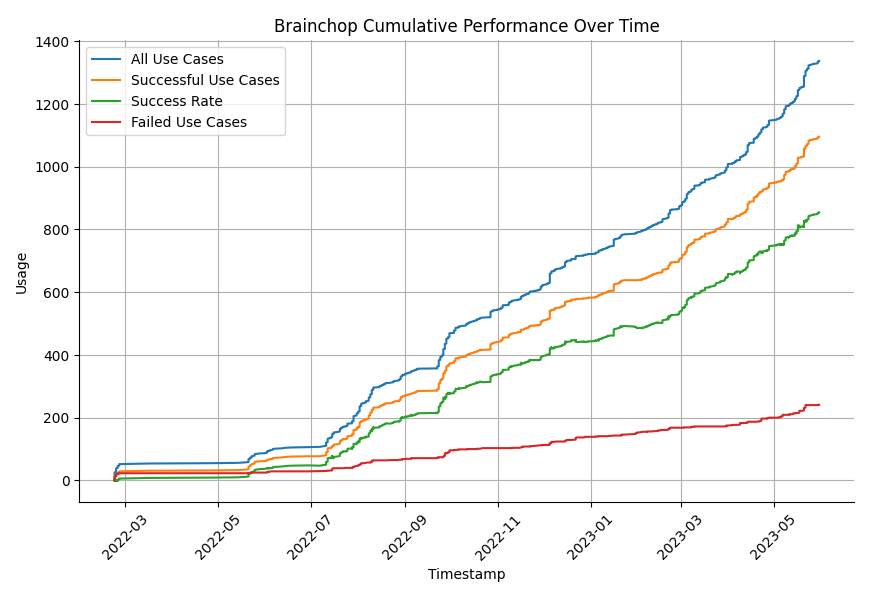}
    \caption{Brainchop cumulative performance over time. Since the tool first version release in May 2022, the disparity between successful hits (orange) and failed hits (red) has been widening, exhibiting a consistent trend of incremental improvement in the success rate (green)  }
    \vspace{-1em} 
\end{figure}

\section*{Acknowledgment}

The authors would like to thank Kevin Wang and Alex Fedorov for discussions and pre-trained MeshNet models

This work was funded by the NIH grant RF1MH121885. Additional support from NIH R01MH123610, R01EB006841 and NSF 2112455.


\vspace{12pt}
\color{red}

\end{document}